\pdfoutput=1
\documentclass{article}

\usepackage{arxiv}
\usepackage[utf8]{inputenc} 
\usepackage[T1]{fontenc}    
\usepackage{url}            
\usepackage{amsfonts}       
\usepackage{graphicx}
\usepackage{doi}
\usepackage[numbers]{natbib}
\usepackage{multicol}
\usepackage{changepage}
\usepackage{float}
\usepackage{listings}
\usepackage{xcolor}
\usepackage{hyperref}       

\lstdefinestyle{python}{
    language=Python,
    backgroundcolor=\color{white},
    basicstyle=\ttfamily\small,
    keywordstyle=\color{blue},
    commentstyle=\color{gray},
    stringstyle=\color{red},
    showstringspaces=false,
    numbers=left,
    numberstyle=\tiny\color{gray},
    stepnumber=1,
    numbersep=10pt,
    frame=single,
    tabsize=4,
    breaklines=true
}

\title{Batching BPE Tokenization Merges}

\date{August 5, 2024}

\author{{Alexander P.~Morgan}\\
	Independent Researcher \\
	\texttt{alex@ecorate.eco} \\
}

\renewcommand{\undertitle}{Lightweight Tokenization Vocabulary Training}

\hypersetup{
pdftitle={Batching BPE Tokenization Merges},
pdfsubject={cm-Data Structures and Algorithms},
pdfauthor={Alexander P. Morgan},
pdfkeywords={tokenization, bpe, byte pair encoding, batch, vocabulary training, llm},
}

\begin{document}
\maketitle

\begin{multicols}{2}
\begin{adjustwidth}{0.025in}{0.025in}
\begin{abstract}
The Byte Pair Encoding algorithm can be safely batched to merge hundreds of pairs of tokens at a time when building up a tokenizer's vocabulary. This technique combined with reducing the memory footprint of text used in vocabulary training make it feasible to train a high quality tokenizer on a basic laptop. This paper presents BatchBPE, an open-source pure Python implementation of these concepts, with the goal of making experimenting with new tokenization strategies more accessible especially in compute- and memory-constrained contexts. BatchBPE's usefulness and malleability are demonstrated through the training of several token vocabularies to explore the batch merging process and experiment with preprocessing a stop word list and ignoring the least common text chunks in a dataset. Resultant encoded lengths of texts are used as a basic evaluation metric.
\end{abstract}
\end{adjustwidth}


\section{Introduction}
Philip Gage introduced the byte pair encoding (BPE) algorithm in 1994 \citep{gage1994new} and its initial purpose was data compression. More recently, BPE has become the primary way text gets tokenized for the training and use of large language models (LLMs) \cite{radford2019language, sennrich2016neural}. It's easy to dismiss this preparatory step as utilitarian or uninteresting, but the way tokenization is handled is critical to an LLM's capabilities, shortcomings, and security issues \cite{land2024fishing}. Tokenization is better thought of as a problem with a passable solution (BPE) with a lot of room for improvement. The fact that every major model release from OpenAI since GPT2 has included significant changes to the tokenization approach used (two doublings of vocabulary size, adding more source document language diversity \cite{hayaseliu2024data}, etc.) demonstrates that this is an aspect of AI research in active development.

I begin this paper with a consideration of tokenization datasets as a distribution of text chunks with an eye to the processing optimizations and useful features this perspective offers. Then I describe how BPE merges can be safely batched to train a tokenization vocabulary considerably faster than when merged one pair at a time. This includes a discussion of the basic issues of batch tokenization and their solutions. A practical implementation of this batch merging tokenization process is given as the open source repository BatchBPE \cite{morgan2024batchbpe} which is a fork of Karpathy's tokenization tutorial repository minbpe \cite{karpathy2024minbpe}. Finally, I apply these learnings by using the BatchBPE tokenizers to explore the impact of preprocessing a stop word list, and discarding the least common text chunks in a dataset. The objective is to see how much can be gleaned from tokenization alone.

\section{Text chunks as a power-law distribution}
It's useful to keep in mind that words, or in our case text chunks, follow a power-law distribution: in any given dataset, a small set of text chunks are very common, and an extremely long tail of others appear only once. This section explains how to take advantage of the uneven nature of text data to speed up vocabulary building, make this process significantly less hardware intensive, and offer special features to potentially improve the tokenization process.

\subsection{Stop word text chunks}
All of the experiments in this paper use the 10B-token sample of the FineWeb-Edu dataset \cite{lozhkov2024finewebedu}. At 27 GB of parquet files (\textasciitilde{}50-60 GB decompressed), this dataset doesn't fit in memory or even in the available hard-disk of my laptop. But we can achieve considerable compression by representing the dataset as a dictionary mapping text chunks

\begin{figure}[H]
    \centering
    \includegraphics[width=\columnwidth]{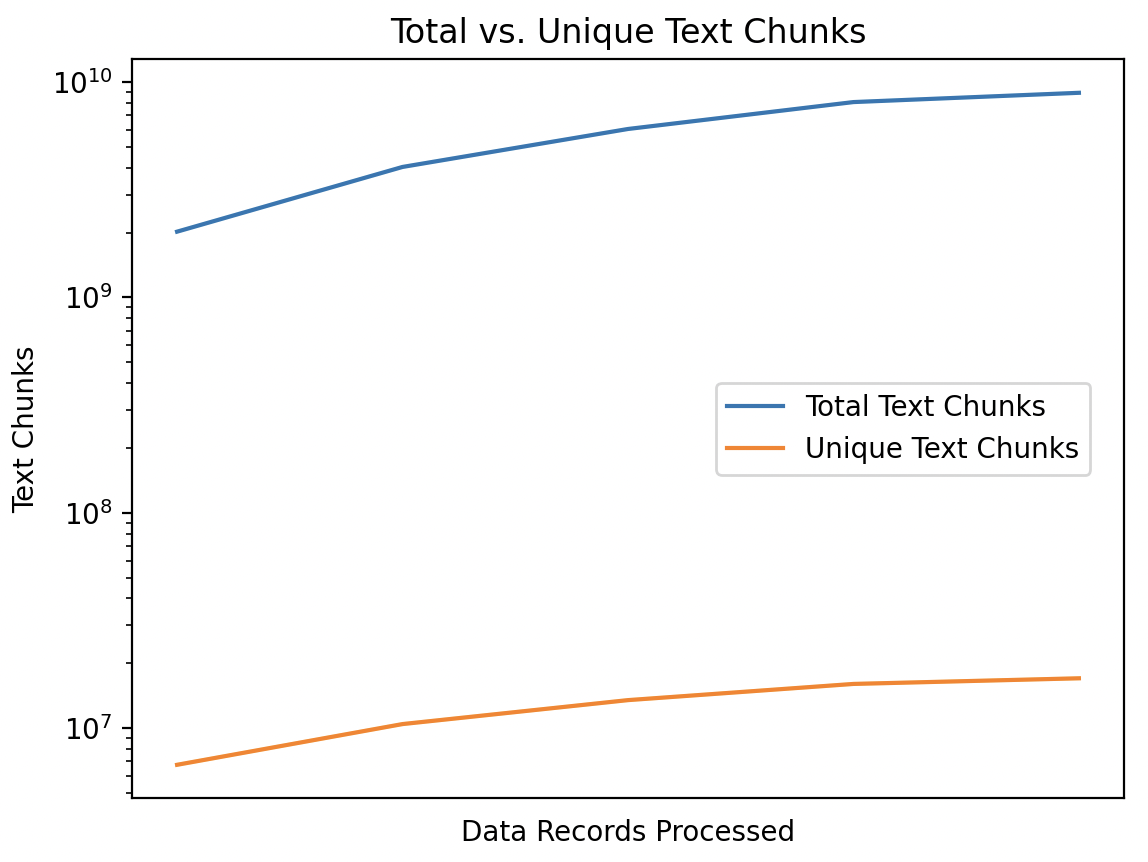}
    \caption{Only a small fraction of total text chunks processed are unique.}
    \label{fig:unique_vs_total}
\end{figure}

to their number of occurrences. This approach saves a lot of runtime by only having to process each text chunk once per batch of token merges, but the most significant advantage is that large datasets can be processed easily without requiring specialized hardware. Figure~\ref{fig:unique_vs_total} shows the unique vs. total text chunks processed when importing the entire dataset. Only 0.19\% of the total text chunks are unique. Later in the vocabulary-building step, each time a text chunk has its token pairs counted, the number of occurrences of that chunk is added to the token pair's tally instead of 1. I also made this consolidated dictionary representation of the FineWeb-Edu 10B sample dataset available as a 2-column csv file \cite{morgan2024finewebedu}.

In total, there are 8.92 billion text chunks in this split of the 10B sample of the FineWeb-Edu dataset (using the GPT4 split pattern), of which just 17.03 million are unique, each appearing 524 times on average. The 100 most common chunks (" the", " in", etc.) represent 45.25\% of the total number of occurrences. This uneven distribution inspired a feature in the BatchBPE tokenizers that allows for the automated preprocessing of the n most common text chunks. This is done by passing the number of stop words you wish to remove as the \verb|stop_list_size| parameter when instantiating a tokenizer. This automated approach (as opposed to passing a literal list of words) makes it easier to handle stop words in a variety of datasets and languages.

When BatchBPE assigns a token to a stop word, that special token is not available during the token merging process. For example, when " the" is preprocessed as a stop word, other words like " theory" won't be able to use that stop word token for " the". Stop word tokens have to be whole-chunk matches. This also means that a stop word token and a token built from the regular token merging process can point to the same literal characters. An experiment showing the relationship between this parameter and encoded text lengths is given in Section~\ref{sec:stop_word_experiment}.

\subsection{Discarding uncommon text chunks}

\begin{figure}[H]
    \centering
    \includegraphics[width=0.85\columnwidth]{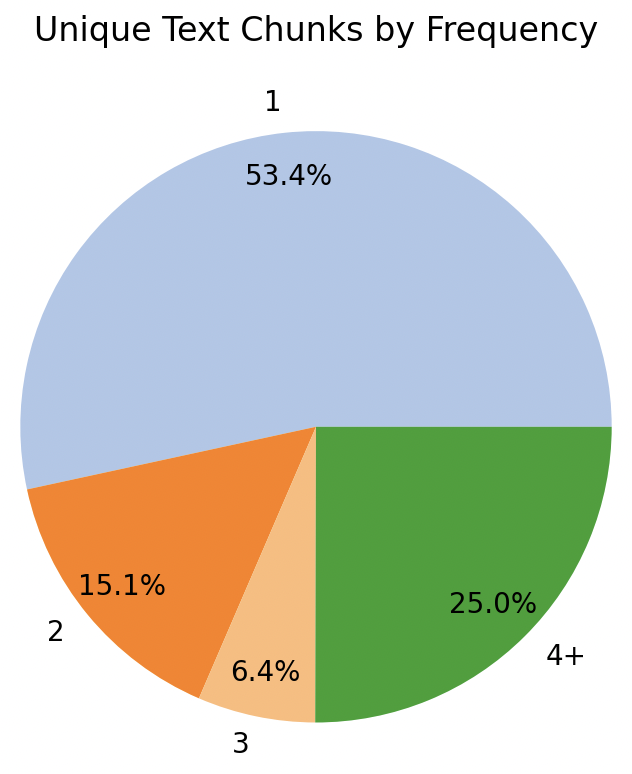}
    \caption{75\% of the unique text chunks in the FineWeb-Edu 10B sample dataset appear fewer than 4 times.}
    \label{fig:text_chunks_by_freq}
\end{figure}

The other end of the text chunk distribution has the opposite characteristic: the majority of unique text chunks appear only once in the data as shown in Figure~\ref{fig:text_chunks_by_freq}. Appearing very few times in so much text is a little suspicious. Are these typos? Even if they correspond to proper words, does processing them improve tokenization outcomes? Inversely, would disregarding these infrequent text chunks improve the process?

To help explore such questions for a given dataset, BatchBPE exposes a \verb|freq_cutoff| parameter that you can pass when instantiating a tokenizer. For example, \verb|BatchTokenizer(freq_cutoff=10)| would instantiate a tokenizer that will remove text chunks that appear fewer than 10 times when the data gets processed. Coincidentally, a \verb|freq_cutoff| setting of 10 resulted in a \textasciitilde{}10X reduction in unique text chunks to process, and by extension in vocabulary building time. Using this parameter can help speed up the vocabulary building process which can be useful if you are trying several different tokenization approaches and have to build a new vocabulary each time. Particularly with a more modest threshold of e.g. \verb|freq_cutoff=2|, using this setting is similar in effect to using a smaller dataset for tokenizer training.

Building a few vocabularies with and without these infrequent text chunks shows that the resulting tokenizations are not identical. In Section~\ref{sec:freq_cutoff_experiment} I show the influence various \verb|freq_cutoff| values have on encoded text length.

After loading the data, the data-structure requirements change. It is necessary to convert the text chunks to bytes

objects and then unpack those to lists of integers. Since lists are not hashable they cannot be used as dictionary keys. So changing the data structure from a dictionary to a list of 2-tuples of text chunks as lists of ints and their frequencies lets us edit the lists of ints in place, saving a lot of copying.

It is possible to combine the steps of merging a batch of token pairs (BatchBPE's \verb|merge_batch|) and recalculating the frequencies of token pairs (BatchBPE's \verb|get_stats|). While this only loops over the text chunks once per batch instead of twice, it only provides a marginal speed improvement (\textasciitilde{}4\%). For simplicity and ease of modification these two steps are kept separate in the \verb|BatchTokenizer| but combined in the \verb|QuickTokenizer|.

\section{Batching token merges}
\label{sec:headings}
The obvious issue with merging tokens in batches is the potential for merges in the same batch to interfere with one another. Accordingly, I first define what safe merges are, then describe the safe batching procedure, and then address some other potential concerns with the approach. Batches are represented as dictionaries mapping (int, int) 2-tuples to a single int new token.

In order to know which two tokens to merge together, we have to scan the entire training dataset, apply a very small amount of changes, then tally all of the token pair frequencies again. Intuitively that is a staggering amount of work to have to do just to merge two tokens. Ignoring the fact that we generally don't merge between split chunks and other details like that, the complexity of the standard BPE vocabulary training process is:

\begin{equation}
    \texttt{Tokens} \times \texttt{VocabSize}
\end{equation}

By batching merges, we're able to reduce this by the average batch size (often 200 or more) to:

\begin{equation}
    \texttt{Tokens} \times \texttt{NumBatches}
\end{equation}

\subsection{Safe merges}
In the construction of a tokenization vocabulary through successive merges, a "safe merge" is one that can be done at the same time as another merge without the two interfering with each other. For example, starting from a merge of the unicode points for the letters "e" and "r" to new token 256, a second merge of the unicode points for "t" and "h" to new token 257 can be considered safe because merging "e" and "r" has no impact on the ability to merge "t" and "h" and vice versa. Thanks to this, it is possible to look for both pairs simultaneously and make the appropriate replacements. They will often both appear in the same chunk of text being tokenized, such as in the word "there", but making one merge does not prevent you from making the other, so they can be grouped into a "safe batch". At a minimum, a safe batch's elements must not interfere with one another. Later in this section I will add to this safe batch definition.

By contrast, the unicode points for "h" and "e" do not represent a safe merge from the same starting point of the "e"-"r" pair. For example, in the word "her", if you first merge "h" and "e", or "e" and "r", either of these merges will make the other unavailable. More explicitly, "her" corresponds to the unicode points (104, 101, 114). Merging "e" and "r", (101, 114), and replacing it with a new token 256, results in (104, 256). This means that "h" and "e", (104, 101), are no longer available together, so merging "e" and "r" has interfered with the ability to merge "h" and "e".

\subsection{Naive safe merges}
The simplest way to get safe batches is to only allow pairs that do not have any tokens in common. We can create a safe batch by considering the token pairs in descending order of frequency. The most common pair will always get included because any batch of size 1 is trivially safe since there are no other pairs to possibly interfere with. Using the example from above of "e" and "r", the batch would begin as \verb|{(101, 114): 256}|. Then we can consider the second-most common pair, "t" and "h", and include it in the batch if it shares no characters in common with any of the others in the batch. That makes the batch \verb|{(101, 114): 256, (116, 104): 257}|. But if the third pair is "h" and "e", this interferes with pair "e"-"r" and also with pair "t"-"h" so it can't be added and we stop searching for more pairs to add to this batch. This naive approach increases batch size significantly and provides an equivalent speed multiplier to the tokenization vocabulary building process.

\subsection{Position-sensitive safe merges}
Instead of checking if the tokens of a new pair have been used anywhere in the existing members of a batch, it is better to check if either token has already appeared in the other position. That is, did the first token in this new pair already appear as the last token in a pair already present in the current batch? Or has the last token in this new pair already appeared as the first token in another batch pair? Building on the batch from before, the "h"-"e" pair causes two conflicts: the "h" is now first, but was previously last in the "t"-"h" pair; similarly the "e" is last but was first in the "e"-"r" pair. By contrast, the "e"-"n" pair would be safe to add to a batch with "e"-"r" and "t"-"h" pairs because, while the "e" does appear again, it does not change position (i.e. "e" is first in both "e"-"r" and "e"-"n") so there is no potential for overlap. The position-sensitive approach roughly doubles the average batch size compared to the naive approach.

\begin{figure*}[t]
    \centering
    \begin{lstlisting}[style=python, caption={Batch merge loop}, label={lst:batch_merge_loop}]
    while merges_remaining > 0:
        seen_first = set()   # tokens seen in the first position in pairs
        seen_last = set()   # tokens seen in the last position in pairs
        pairs_to_merge = {}
        stats = get_stats(ids)   # update token-pair frequencies
        num_pairs_to_search = min(merges_remaining//cap_divisor, len(vocab), max_batch_size) or 1
        top_pairs = heapq.nlargest(num_pairs_to_search, stats, key=stats.get)
        for first, last in top_pairs:  # pairs are (first, last) tuples
            if first in seen_last or last in seen_first:   # unsafe merge
                seen_first.add(first)
                seen_last.add(last)
                continue # skip this pair but keep looking for safe merges
            seen_first.add(first)
            seen_last.add(last)
            pairs_to_merge[(first, last)] = curr_vocab_size
            vocab[curr_vocab_size] = vocab[first] + vocab[last]
            curr_vocab_size += 1
        merges_remaining -= len(pairs_to_merge)
        merges.update(pairs_to_merge)  # save the merges
        batch_count += 1
        if merges_remaining:   # no need to merge last batch
            merge_batch(ids, pairs_to_merge)
    \end{lstlisting}
\end{figure*}

\subsection{Continue searching}
When encountering a token pair that is not safe to include in a batch, instead of stopping there we can omit that unsafe pair from the batch but keep searching for others. In continuing to search for safe merges, we also need need to ensure that further token pairs that we consider don't conflict with any of the skipped pairs. Simply adding the tokens of the pairs that get skipped to the seen-token sets accomplishes this. As more and more token pairs get added to these seen-token sets, it becomes increasingly difficult to find a safe merge, however, the approach still greatly increases the average batch size. A Python implementation of this loop is given in Code Block~\ref{lst:batch_merge_loop}.

With this new approach in mind, we should update our definition of a safe batch. A safe batch is one where none of the member pairs overlap with one another and each member also does not overlap with any more frequent pairs that were skipped in the building up of the batch.

\subsection{Batch merging issues and solutions}
There are three issues to be aware of when batch merging token pairs, but they are easily remedied. The first is that you should never consider merges that are taken from outside of the target vocabulary size. For example, if your target vocabulary size is 50,304, then you should not consider the 50,049\textsuperscript{th} most-common token pair. That's because we start with a vocabulary of size 256 (for the stock unicode points) so we can only do 50,048. This means that we have to keep track of the number of merges remaining and use this as a ceiling on the number of token pairs to consider at each batch.

Similarly, we shouldn't consider token merges from the very last pairs in our ordered list of remaining pairs to merge. When a higher frequency token pair is merged, this will often split less common token pairs into multiple pairs. For example, when merging "t" and "h" into 257, the occurrences of the "h"-"e" pair will get replaced by some that are now 257-"e" and others that are still "h"-"e". The number of possible pairs to merge always increases (unless the vocabulary is severely oversized for the training text) but the number of occurrences of a given pair can only decrease before it gets merged because of the splitting described above. So the frequency ranking of unsafe merges (the ones we skip when making a safe batch) is fluid. BatchBPE addresses both of these issues by implementing a \verb|cap_divisor| parameter. It caps the number of pairs searched for inclusion in the current batch to the number of remaining merges divided by the \verb|cap_divisor| (default 2). Using floor division this can be zero so we also have to enforce that at least one pair is searched.

A different issue manifests itself at the beginning of vocabulary building. The vocabulary size starts at 256 (for the 256 stock bytes object values), but when making the first batch, no splitting of token pairs has occurred yet, so we need a different early limit on the number of pairs to consider. Otherwise we might include rare pairs like "z"-"q" which do occur but not enough to be included in a vocabulary when lots of more frequent pairs have created more pairs through the splitting process. BatchBPE addresses this by also limiting the number of pairs searched to be no greater than the current vocabulary size. This limit grows throughout the training process, in contrast to the \verb|cap_divisor| ceiling that starts high and continually lowers. The combination of these two limits means that the largest batches in a training run are generally found when about one third of the merges have been made.

BatchBPE tokenizers also expose a \verb|max_batch_size| parameter which imposes a hard limit on the number of pairs to search for safe merges. Setting it to 1 would effectively build the vocabulary serially rather than in batches. These safeguards for the batch tokenization process are combined in the simple formula given in line 6 of Code Block~\ref{lst:batch_merge_loop}.

\subsection{Repeated token pairs}
One concern remains. When a pair consists of one token occurring twice, it is possible that immediately after making this merge, the next most common pair would be that new token appearing twice in a row. Such situations can occur, for example, with spaces when processing code-heavy datasets. In fact, the first two merges of the GPT4o vocabulary (tokens 256 and 257) are merges of two single spaces, and then two double-spaces (four spaces in total) respectively \cite{duong2024tiktokenizer}.

The batch merging process would not have even considered this second merge in the first batch. Analyzing tokenizer vocabularies can indirectly reveal what kind of dataset was used for training \cite{hayaseliu2024data}. Similarly, by analyzing the order of a tokenizer's vocabulary we can deduce whether it was built up serially or in batches. It appears that OpenAI did not use a batch merging process to build its most recent tokenizer vocabulary, at least not at the beginning of vocabulary building. As a side note, OpenAI also stopped splitting text at apostrophes starting with GPT4o.

Land and Bartolo showed that you can analyze a tokenization vocabulary to find unused, inaccessible, or undertrained tokens \cite{land2024fishing}. Beyond the first two merges for double and quadruple spaces mentioned above, every number of consecutive spaces from 1 to 79 inclusive has its own dedicated token in OpenAI's GPT4o vocabulary \cite{duong2024tiktokenizer}. So it seems that little consideration was given to minimizing that token vocabulary size.

 In practice, repeated-token pairs are not problematic for the batch merging process. That's because the first pair of two single spaces already protects the later pair of two double spaces since no other pairs that include spaces will be included in the first batch. So the quadruple-space pair will simply be merged in the next batch.

The only case where this can possibly cause a change in the tokenization is if this merge happens just before the final batch. Due to BatchBPE's limit of searching at most half of the remaining merges, and the fact that there is a great variety of pairs to count near the end of vocabulary building, BatchBPE will often do batches of sizes that descend powers of two in the last few batches (limited in size by the \verb|cap_divisor| parameter described above), e.g. 8, 4, 2, 1, 1. Using repeated a's instead of spaces for readability, imagine that the third-to-last batch (of size 2), the most common pair was \verb|{("a", "a"): 50300, ("z", "y"): 50301}|. Imagine further that \verb|{(50300, 50300): 50302}| was the second-to-last batch, and \verb|{(50302, 50302): 50303}| was the final batch. The vocabulary building would be over but it would still technically be possible for the next most-common pair to be \verb|(50303, 50303)| and for this pair to still be more common than the next-most common pair merged in the third-to-last batch, \verb|("z", "y")| in this example. In that highly improbably scenario, the batch merging process would have only this tiny difference in its vocabulary compared to the standard approach, so this is a non-issue.

Somewhat related to this is the issue of over-counting pair occurrences when dealing with repeated character strings. BatchBPE's \verb|BatchTokenizer| handles that issue in the way it counts pairs, so, for example, "aaaaa" counts as two occurrences of the "a"-"a" pair (since you can only make that merge twice) instead of four. This approach been presented as prioritizing potential for compression over raw token-pair count \citep{zouharetal2023formal}. In contrast, BatchBPE's \verb|QuickTokenizer| overcounts these pairs and also integrates a few other changes to offer a slightly faster tokenizer alternative and above all to demonstrate how to make different tokenizer subclasses to accommodate different tokenization objectives.

\section{Tokenization experiments}
\label{sec:experiment}
With the different features of BatchBPE and the batch token merging process explained, this section offers a demonstration of the kinds of insights that can be gained through tokenization alone. I use the impact on the tokenized length of a text as a simple way to check the usefulness of different techniques. Measuring the validation loss of a trained LLM that uses these token vocabularies would be a more significant metric for determining whether any of these approaches are beneficial to a given model, however, no LLMs are trained in this study. The techniques vary in their usefulness for larger or smaller models and different training language distributions. Accordingly, I limit the remainder of this study to general observations about the approaches introduced above. I used the FineWeb-Edu 10B sample dataset \cite{lozhkov2024finewebedu} to train all of the tokenization vocabularies presented in this paper to 50,304 token vocabularies.

\subsection{Disambiguating stop words as prefixes}
\label{sec:stop_word_experiment}
In a vocabulary I built with the \verb|BatchTokenizer|, I included " in" as a preprocessed stop word, among others. As a standalone word " in" relates to location or membership, but as a prefix it usually indicates negation, which is roughly the opposite meaning. While devoting multiple tokens to the same byte combinations reduces the variety of byte combinations you can represent with a given vocabulary size, it can disambiguate the most common cases of byte-level homographs. The catch is that once a token merge does happen, the new token will not inherit any of the meaning that the LLM will later learn of the two earlier tokens. One promising way to minimize ambiguities

\begin{figure}[H]
    \centering
    \includegraphics[width=0.75\columnwidth]{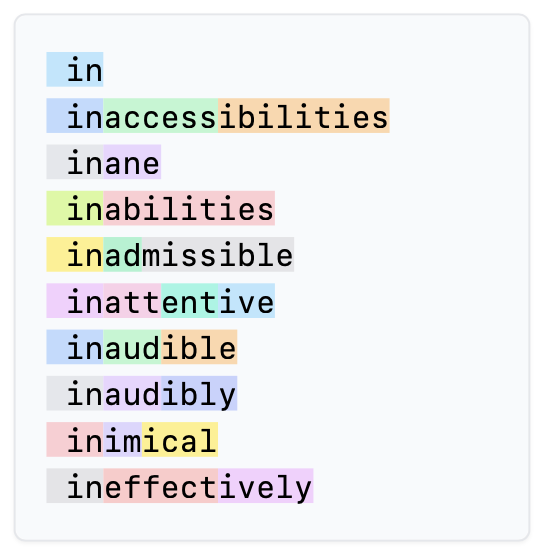}
    \caption{Several longer words in GPT2's token vocabulary begin with the token for " in".}
    \label{fig:in_words}
\end{figure}

is to search for stop words that also serve as prefixes of other text chunks in a dataset. This is highly dependent on the dataset, vocabulary size, and split pattern used. To continue with the " in" example, in the GPT2 vocabulary (\textasciitilde{}50k tokens), " ineffective" is represented by a single token, 23,693. Having its own token, " ineffective" doesn't encapsulate any of the probabilities of sequencing that "~in" (token 287) has. However, there are several other text chunks that start with " in" that don't coalesce into a single token. The color-coding scheme of the Tiktokenizer web app \cite{duong2024tiktokenizer} is a clear visual way to convey the token breakdowns of text, and this tool is used in Figure~\ref{fig:text_chunks_by_freq} to display the GPT2 tokenizer's breakdown of several multi-token words starting with " in". 

The number of stop words preprocessed has a small but mostly adverse impact on the average encoded length of text. Using the \verb|BatchTokenizer|, I trained 10 vocabularies iterating the \verb|stop_list_size| parameter from 0 to 200 at increasingly large intervals. Change in encoded length of 1 GB of input text varied only slightly, between

\begin{figure}[H]
    \centering
    \includegraphics[width=0.9\columnwidth]{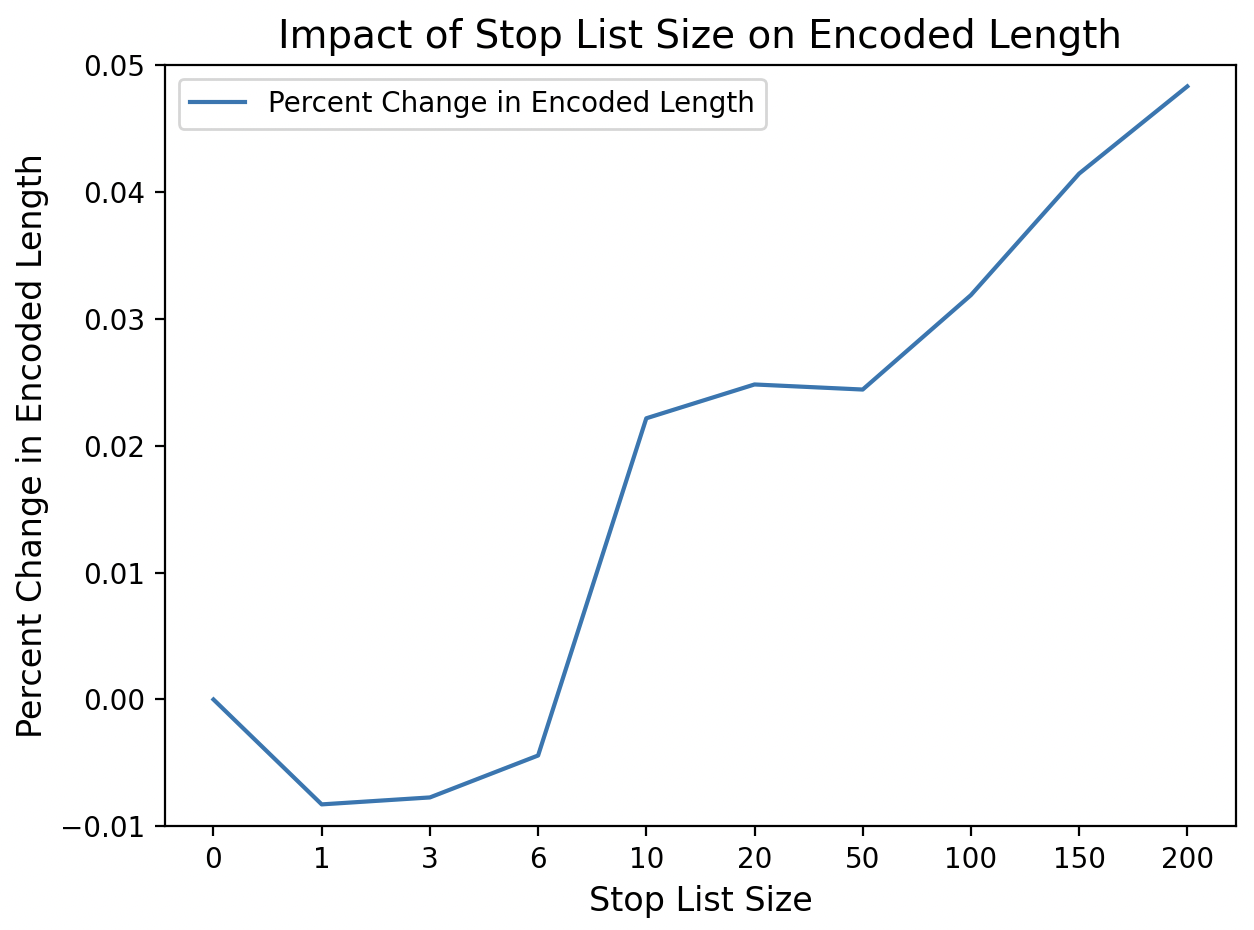}
    \caption{Preprocessing a few stop words has a marginal impact on encoded length.}
    \label{fig:stop_list_size_vs_encoded_length}
\end{figure}

-0.0083\% and 0.0483\% as shown in Figure~\ref{fig:stop_list_size_vs_encoded_length} (all percentages given in this paper are out of 100, not 0-1). Some configurations even decreased the encoded length, presumably because preprocessing a few of the most common words allowed the regular BPE process to begin on a more efficient path for the remainder of the token pairs.

\subsection{Filtering rare text chunks}
\label{sec:freq_cutoff_experiment}
\begin{figure}[H]
    \centering
    \includegraphics[width=0.9\columnwidth]{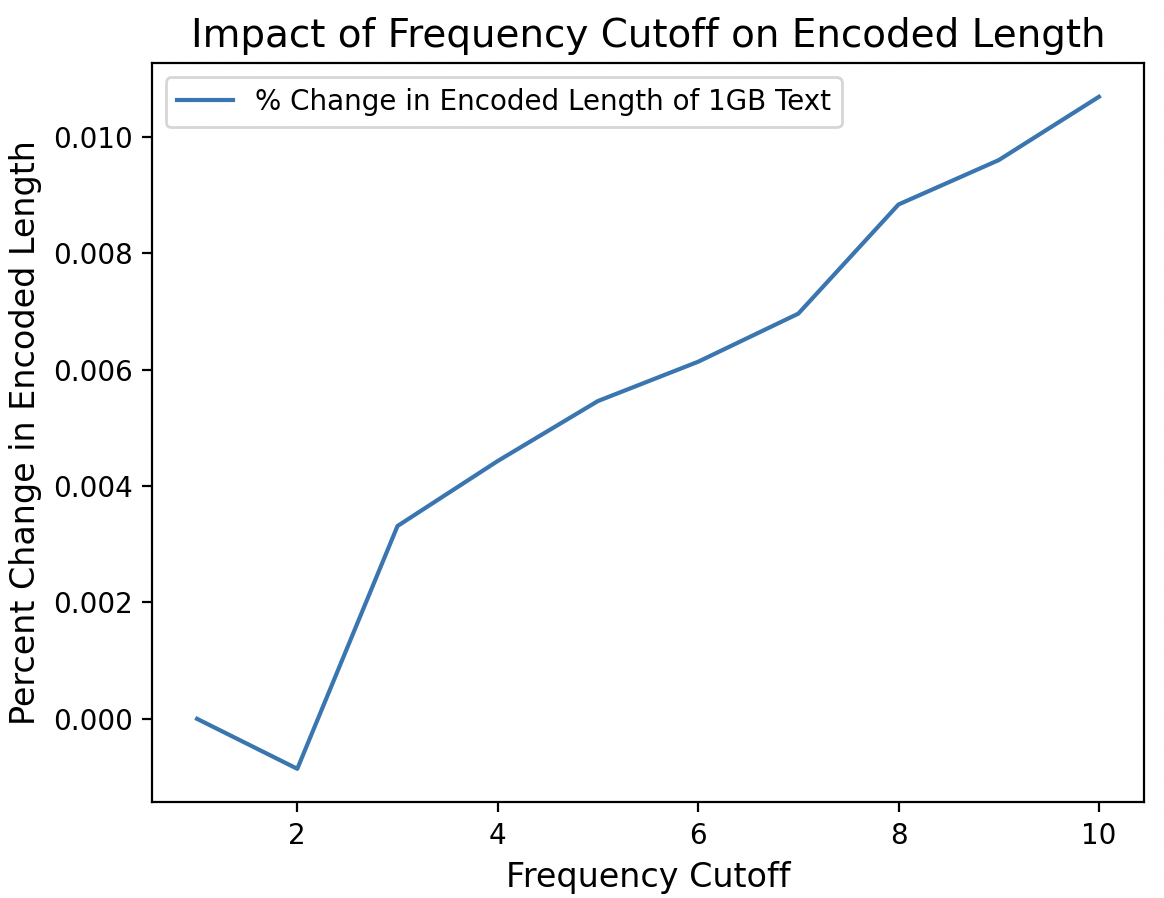}
    \caption{Removing rare words from a dataset generally has a slight adverse impact on encoded text length.}
    \label{fig:freq_cutoff_chart}
\end{figure}

Building a vocabulary with the \verb|freq_cutoff| set above 1 also has a small influence on encoded text lengths. Figure~\ref{fig:freq_cutoff_chart} plots the percent change in encoded-text length for vocabularies of 50,304 tokens with \verb|freq_cutoff| settings from 1 (i.e. no cutoff) to 10 inclusive. While the slight increase for most values is generally undesirable, it can be useful to train tokenizers with the \verb|freq_cutoff| as high as 10 to quickly experiment with one's overall approach. Once a final approach and combination of settings have been decided on, you can retrain the final vocabulary with no \verb|freq_cutoff|. Put another way, I have found that setting \verb|freq_cutoff=10| is a good indicator of a vocabulary built without frequency filtering. Using \verb|freq_cutoff=2| even marginally lowered the encoded text length in this experiment. Since the FineWeb-Edu dataset I used for training is highly refined and reliable, it's easy to imagine that other noisier datasets might benefit more from using the \verb|freq_cutoff| to cull uncommon text chunks.

\section{Conclusion}
In this paper I defined what safe merges are with respect to batching the token pair merging process in the BPE algorithm, and reviewed concerns with this approach along with recommended solutions. I also introduced a new open source tokenizer, BatchBPE \cite{morgan2024batchbpe}, which uses this batch merging technique and went into detail about the tool's features that allow for the automated preprocessing of the most common stop words, and the removal of rare words from a dataset before training.

The two critical enablers of tokenization training used in BatchBPE are condensing the training dataset into a dictionary mapping text chunk strings to their frequencies, and merging token pairs in batches. Without the first, it is impractical to update token pair counts on basic hardware. I made a text chunk to count mapping of the FineWeb-Edu 10B sample dataset \cite{lozhkov2024finewebedu} available as a csv file \cite{morgan2024finewebedu} that is over 100 times smaller than the original. Other tokenizers \citep{openai2022tiktoken, huggingface2024tokenizers, kudorichardson2018sentencepiece, forsythe2023tokenmonster, forsythe2023capcode} written in more performant programming languages certainly encode and decode text and tokens faster than BatchBPE. However, the latter's strength is in the ease and speed it allows you to train a new vocabulary and experiment with tokenization approaches. I highly recommend these two core enabling techniques of batching token merges and representing the data to process as a dictionary for other tokenizers. In particular, I think that batching token pair merges will become increasingly important as vocabulary sizes grow.

It's easy to imagine further developments to the BPE approach for tokenization or even to question it as the primary approach \cite{bostromdurrett2020bpe}. Beyond stop words, it would also be possible to implement processing for multiple text chunk tokens, e.g. a single token for common phrases such as "~of~the", "~how~much", or ",~however,". Doing this in a way similar to how BatchBPE preprocesses stop words would work, though would add some complexity. A promising alternative approach could be to merge tokens in two phases, somewhat differently than Yang \cite{yang2024rethinking}. The first phase would be the now-standard approach of merging token pairs within text chunks according to a split pattern. After enough of these kinds of merges, a second phase could take over that merges the most common adjacent token pairs even if they come from different text chunks. This could be an elegant way to combine byte-, word-, and phrase-level tokenization approaches.

\end{multicols}

\bibliographystyle{unsrtnat}

\end{document}